%% file: main_v3_arxiv.tex
\documentclass[10pt,twocolumn,letterpaper]{article}

\usepackage[pagenumbers]{cvpr} 

\input{preamble}
\definecolor{cvprblue}{rgb}{0.21,0.49,0.74}
\usepackage[pagebackref,breaklinks,colorlinks,allcolors=cvprblue]{hyperref}

\usepackage[utf8]{inputenc} 
\usepackage[T1]{fontenc}    
\usepackage{hyperref}       
\usepackage{url}            
\usepackage{booktabs}       
\usepackage{amsfonts}       
\usepackage{nicefrac}       
\usepackage{microtype}      
\usepackage[table,dvipsnames]{xcolor}

\usepackage{multirow}
\usepackage{units}
\usepackage{pifont}
\usepackage{wrapfig}
\definecolor{darkgreen}{RGB}{0,120,0}
\definecolor{softgreen}{RGB}{0,100,0}
\usepackage{amsmath}
\usepackage{amssymb}
\usepackage{mathtools}
\usepackage{amsthm}
\usepackage{enumitem}
\usepackage{bm}

\makeatletter
\def\thanks#1{\protected@xdef\@thanks{\@thanks
        \protect\footnotetext{#1}}}
\makeatother

\title{LongVie~2: Multimodal Controllable Ultra-Long Video World Model}

\author{
Jianxiong Gao$^1$, Zhaoxi Chen$^3$, Xian Liu$^4$, Junhao Zhuang$^5$, Chengming Xu$^1$ \\
Jianfeng Feng$^1$, Yu Qiao$^6$, Yanwei Fu$^{1,\dagger}$, Chenyang Si$^{2,\dagger}$, Ziwei Liu$^{3,\dagger}$
\thanks{$\dagger$: Co-corresponding authors.}\\[2pt]
$^1$FDU,\quad
$^2$NJU,\quad
$^3$NTU,\quad
$^4$NVIDIA,\quad
$^5$THU,\quad
$^6$Shanghai AI Laboratory \\[2pt]
\textcolor{violet}{\textit{https://vchitect.github.io/LongVie2-project/}}\\
}

\begin{document}

\twocolumn[{
\renewcommand\twocolumn[1][]{#1}
\maketitle

\begin{center}
\vskip -0.05in
\includegraphics[width=1.0\linewidth]{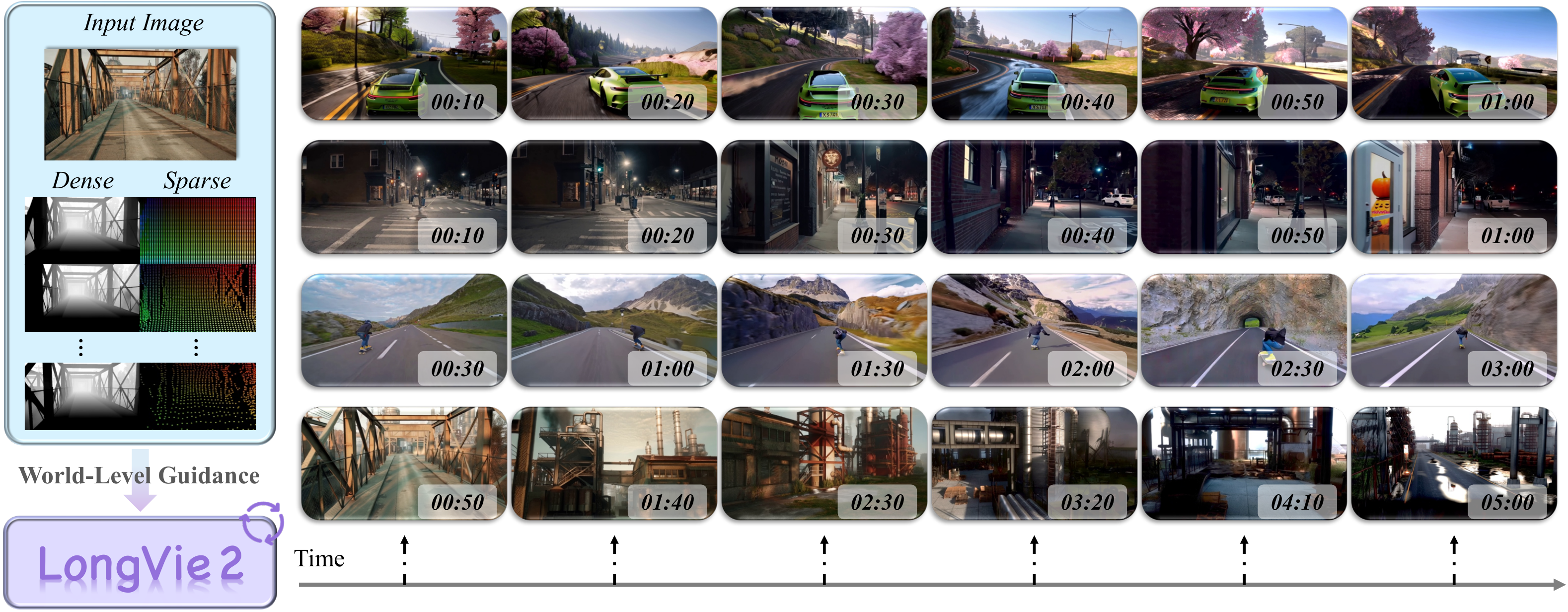}
\vskip -0.05in
\captionof{figure}{
\label{fig:teaser}
\textbf{LongVie~2} is a controllable ultra-long video world model that autoregressively generates videos lasting up to 3–5 minutes.
It is driven by world-level guidance integrating both dense and sparse control signals, trained with a degradation-aware strategy to bridge the gap between training and long-term inference, and enhanced with history-context modeling to maintain long-term temporal consistency.
}
\end{center}
}]

\begin{abstract}

Building video world models upon pretrained video generation systems represents an important yet challenging step toward general spatiotemporal intelligence.
A world model should possess three essential properties: controllability, long-term visual quality, and temporal consistency.
To this end, we take a progressive approach—first enhancing controllability and then extending toward long-term, high-quality generation.
We present \textbf{LongVie~2}, an end-to-end autoregressive framework trained in three stages:
\textbf{(1)} Multi-modal guidance, which integrates dense and sparse control signals to provide implicit world-level supervision and \textbf{improve controllability};
\textbf{(2)} Degradation-aware training on the input frame, bridging the gap between training and long-term inference to maintain \textbf{high visual quality}; and
\textbf{(3)} History-context guidance, which aligns contextual information across adjacent clips to \textbf{ensure temporal consistency}.
We further introduce \textbf{LongVGenBench}, a comprehensive benchmark comprising 100 high-resolution one-minute videos covering diverse real-world and synthetic environments.
Extensive experiments demonstrate that LongVie~2 achieves state-of-the-art performance in long-range controllability, temporal coherence, and visual fidelity, and supports continuous video generation lasting up to \textbf{five} minutes, marking a significant step toward unified video world modeling.
\end{abstract}

\section{Introduction}

Recent breakthroughs in video generation have been largely driven by the availability of large-scale datasets and the rapid development of diffusion-based architectures~\cite{ho2020DDPM, rombach2021highresolution, opensora, meta_moviegen, dataverse}. These advances have enabled powerful models such as CogVideoX~\cite{yang2024cogvideox}, HunyuanVideo~\cite{kong2024hunyuanvideo}, Kling~\cite{kling}, Sora~\cite{sora}, and Wan2.1~\cite{wan2025} to synthesize photorealistic and semantically rich videos directly from text prompts, demonstrating that diffusion models can effectively capture both visual dynamics and high-level semantic structures.

\begin{figure}[t]
    \centering
    \includegraphics[width=0.99\linewidth]{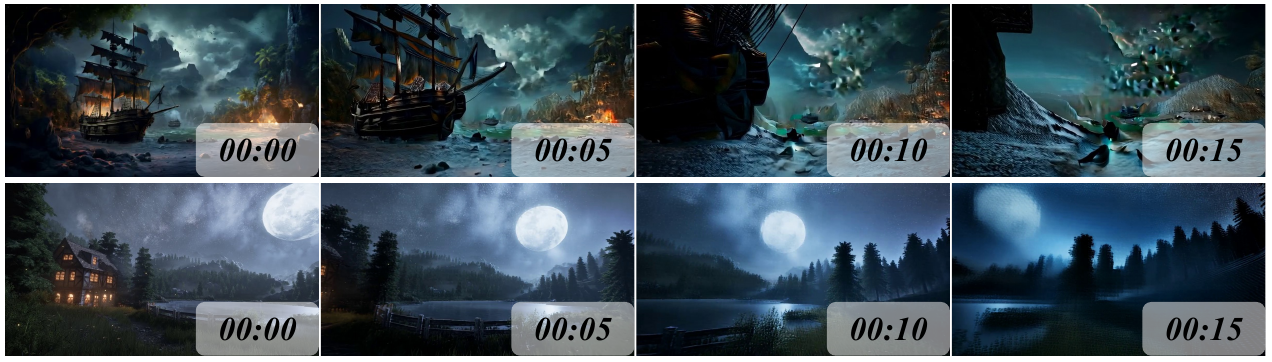}
    \vskip -0.05in
    \caption{
    \textbf{Unstable long-term generation.} 
    As the generated duration increases, current video world models gradually lose controllability, visual fidelity, and temporal consistency.
    }
    \label{fig:intro_problem}
    \vskip -0.05in
\end{figure}

Building upon this foundation, research has recently shifted from merely generating visually appealing clips to modeling the underlying dynamics of the physical world. This shift has led to the emergence of video-based world models that simulate realistic environments and interactions, exemplified by HunyuanGameCraft~\cite{li2025hunyuangamecraft}, Yume~\cite{mao2025yume}, and Matrix-Game~\cite{he2025matrix}. These models incorporate controllable elements—such as camera motion or agent actions—to emulate how the world evolves over time.
However, despite these advances, current world models face two major challenges. First, their controllability remains limited, often restricted to low-level or localized adjustments rather than global, semantic-level control over the environment. 
Second, their temporal scalability is fragile—when extended to long horizons (\eg beyond one minute), these models often exhibit visual degradation and temporal drift, as shown in Fig.~\ref{fig:intro_problem}, where visual quality collapses over time.

Toward the next generation of video world models, we seek to unify two complementary objectives—fine-grained controllability and long-term coherence—within a single framework. An ideal world model should not only generate visually and temporally consistent videos but also respond coherently to structured inputs and interactions. In essence, it should achieve:
(1) \textbf{Comprehensive controllability}, ensuring that the entire evolving scene remains semantically interpretable and manipulable across time;
(2) \textbf{Long-term fidelity}, maintaining visual quality over extended sequences; and
(3) \textbf{Long-context consistency}, preserving stable, realistic dynamics aligned with real-world behavior.

In this paper, we investigate how to construct video world models by extending pretrained video diffusion backbones into controllable long-horizon generators. Introducing controllability into these pretrained architectures is relatively straightforward, as most short-clip diffusion models already support modular conditioning through components such as LoRA~\cite{hu2021loralowrankadaptationlarge}, ControlNet~\cite{controlnet}, or similar plug-in adapters.
Leveraging this property, we adopt a control-first, then long strategy: we first enhance controllability within short video generation and then progressively extend the temporal horizon to achieve long-range coherence. 

Accordingly, we extend controllable short-video generators into long-horizon, temporally coherent frameworks, marking an essential step toward general-purpose video world modeling. To this end, we propose LongVie~2, a controllable long video generation framework built upon an autoregressive paradigm. LongVie~2 is trained through three progressive stages that jointly enhance controllability, temporal consistency, and visual fidelity:
\textbf{1) Multi-modal guidance} integrates dense (\eg, depth maps) and sparse (\eg, keypoints) control signals under balanced training, providing complementary structural and semantic constraints that offer implicit world-level guidance.
\textbf{2) Degradation-aware} training introduces controlled degradations to guided frames, effectively bridging the gap between training and long-horizon inference.
\textbf{3) History context guidance} leverages preceding frames as contextual inputs to maintain long-range consistency across adjacent clips.
Together, these stages enable LongVie~2 to generate minute-long controllable videos with strong temporal coherence and high visual fidelity, marking a crucial step toward unified and scalable video world models.

To evaluate the effectiveness of \textbf{LongVie~2}, we construct \textbf{LongVGenBench}, a benchmark consisting of 100 high-quality videos, each lasting at least one minute. The dataset covers a wide range of scenarios, including both real-world environments and game-based scenes, spanning day and night as well as indoor and outdoor settings, ensuring rich variations in content, structure, and motion dynamics. We compare LongVie~2 and baselines on LongVGenBench in terms of long-term visual quality and temporal consistency. Experimental results demonstrate that LongVie~2 achieves state-of-the-art performance in controllable long video generation, producing videos both temporally coherent and visually realistic.

\begin{figure*}
\begin{center}
\includegraphics[width=\linewidth]{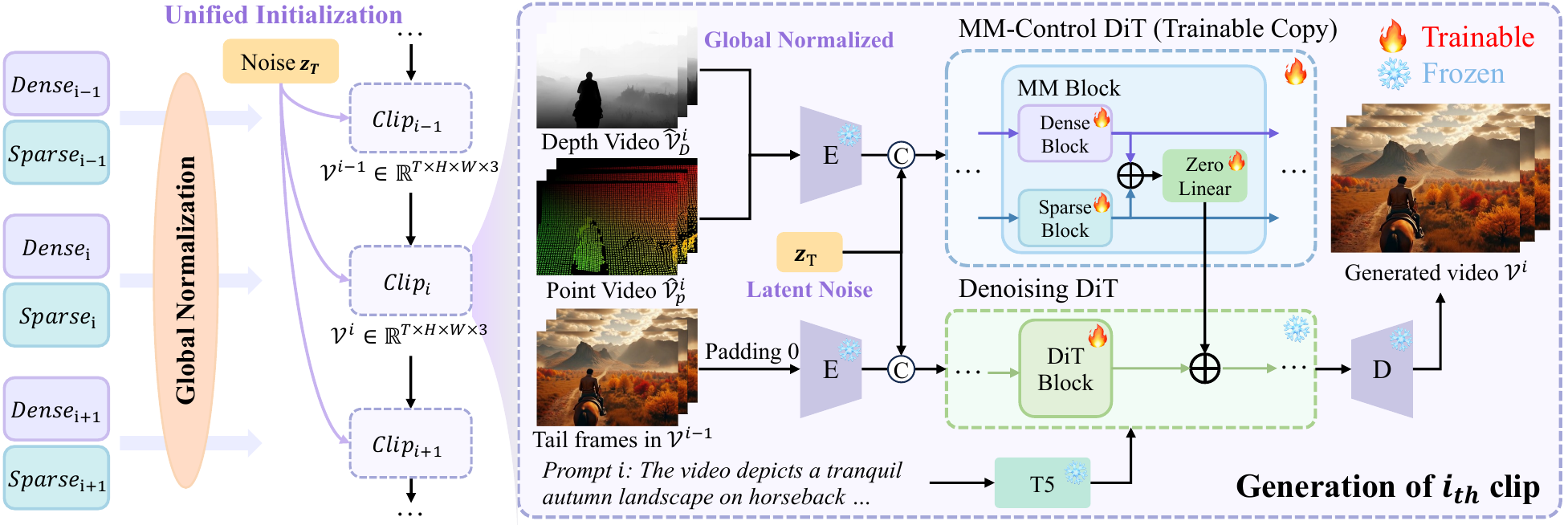}
\vskip -0.1in
\caption{
\textbf{Framework of LongVie~2.}
\textbf{LongVie~2} serves as a controllable video world model that integrates both dense and sparse control signals to provide world-level guidance for enhanced controllability. A degradation-aware training strategy improves long-term visual quality, while tail frames from preceding clips are incorporated as historical context to maintain temporal consistency over extended durations.
}
\label{fig:framework}
\end{center}
\vskip -0.2in
\end{figure*}

Our main contributions are summarized as follows:

\begin{itemize}
\item We present \textbf{LongVie~2}, a controllable long video generation framework that extends pretrained diffusion backbones toward video world modeling.
\item We design a \textbf{three-stage progressive training scheme} that integrates multi-modal guidance, degradation-aware learning, and history context guidance to enhance controllability, temporal coherence, and visual fidelity, aligning with essential capabilities required for video world models.
\item We establish \textbf{LongVGenBench}, a benchmark of 100 one-minute videos for evaluating controllability, consistency, and perceptual quality in long video generation.
\end{itemize}

\section{Methodology}

\textbf{Preliminary: Diffusion Models.}
Diffusion models are powerful video world learners that synthesize realistic videos by progressively removing noise. So, the model $\epsilon_\theta$ learns to undo a process that adds noise to data. During training, it is trained to predict the noise added at each step:
\begin{align}
    \label{eqn_loss}
    \mathcal{L} = \mathbb{E}_{\boldsymbol{x},\epsilon \sim \mathcal{N}(0,1),t} \Big [  \lVert \epsilon - \epsilon_\theta(\boldsymbol{x}_{t}, t) \rVert^2_2 \Big ],
\end{align}
where $\boldsymbol{x}_t$, $\epsilon$, $\mathbf{x}_0$ denote the noisy data at time $t$,  the true noise, and the original data, respectively.
To save computing power, Latent Diffusion Models~\cite{rombach2021highresolution} work in a smaller, compressed space. An autoencoder turns $\boldsymbol{x}$ into a latent code $\boldsymbol{z} = \mathcal{E}(\boldsymbol{x})$, and the model learns to predict noise in $\boldsymbol{z}_t$.

\noindent \textbf{Overview.}
To address the aforementioned challenges and advance toward a unified video world model, we propose LongVie~2, an end-to-end framework trained through three progressive stages, as illustrated in Fig.~\ref{fig:framework}.
In Stage I, we inject multi-modal control signals into the model following the standard ControlNet paradigm, enabling implicit world-level guidance that enhances controllability.
In Stage II, we introduce a first-frame degradation strategy that simulates long-form deterioration during training, effectively mitigating visual degradation in long-horizon generation.
In Stage III, we incorporate historical frames as RGB guidance, allowing the model to leverage temporal context and maintain coherence across adjacent clips.

\begin{figure*}
\begin{center}
\includegraphics[width=\linewidth]{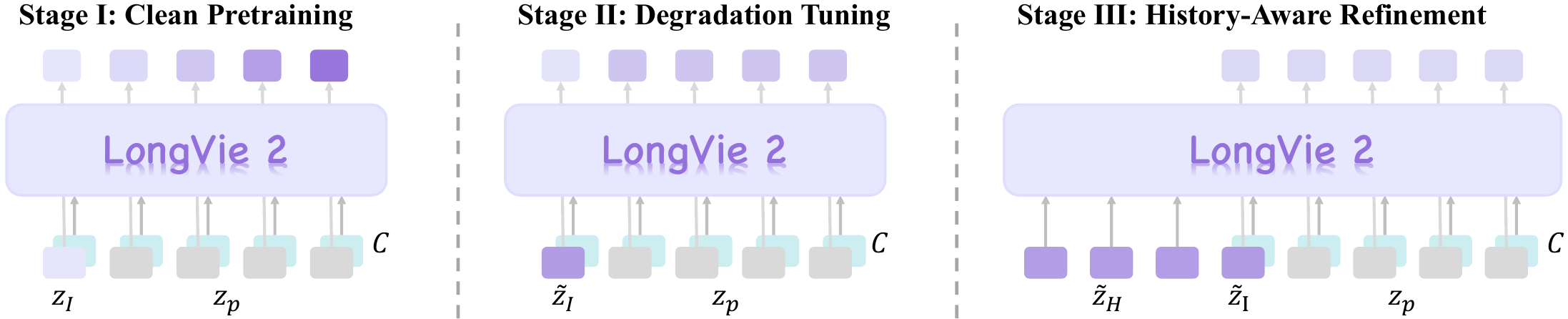}
\vskip -0.05in
\caption{\textbf{Training pipeline of LongVie~2.} We first train the model using a standard ControlNet-based pipeline. In the second stage, we introduce degradation to the first frame to bridge the domain gap between ground truth and generated frames. Finally, to ensure temporal consistency, we incorporate historical frame information.}
\label{fig:training_pipeline}
\end{center}
\vskip -0.2in
\end{figure*}

\subsection{Multi-Modal Control Injection}
Specifically, we adopt depth maps as dense control signals and point maps as sparse control signals, leveraging the detailed structural information provided by depth and the high-level semantic cues captured by point trajectories to form an implicit world representation.
To construct the point map sequences, we follow the procedure in DAS~\cite{gu2025das}, where a set of keypoints is tracked across frames and colorized according to their corresponding depth values.

Inspired by the ControlNet~\cite{controlnet} architecture, we build our Multi-Modal Control DiT by duplicating the initial 12 layers of the pre-trained Wan DiT and incorporating multi-modal conditioning inputs while keeping the base model frozen. Specifically, we freeze the parameters $\theta$ of the original DiT blocks and construct two trainable branches, each corresponding to a distinct control modality: dense and sparse. These branches, denoted as $\mathcal{F}_\text{D}(\cdot;\theta_\text{D})$ and $\mathcal{F}_\text{P}(\cdot;\theta_\text{S})$ , are lightweight copies of the original DiT layers with parameters $\theta_\text{D}$ and $\theta_\text{P}$ , respectively. Each processes the corresponding encoded control signal $\boldsymbol{c}_\text{D}$ or $\boldsymbol{c}_\text{P}$. To integrate the control branches into the base generation path, we adopt zero-initialized linear layers $\phi^l$.
These layers allow the control signals to be injected additively into the main generation stream without affecting the model’s initial behavior, as they output zero at the start of training. The overall computation in the $l$-th controlled DiT block is defined as:
\begin{align}
    \label{eqn_fuse}
    \boldsymbol{z}^l = \mathcal{F}^l( \boldsymbol{z}^{l-1}) + \phi^l(\mathcal{F}_\text{D}^l(\boldsymbol{c}^{l-1}_\text{D}) + \mathcal{F}_\text{S}^l(\boldsymbol{c}^{l-1}_\text{S}) ),
\end{align}
where $\mathcal{F}^l$ is the frozen base DiT block, and $\mathcal{F}_\text{D}^l$, $\mathcal{F}_\text{P}^l$ are the control-specific sub-blocks.

While multi-modal control offers the potential for richer and more accurate video generation, simply combining dense and sparse control signals does not guarantee improved performance. In practice, we observe that dense signals tend to dominate the generation process, often suppressing the semantic and motion-level guidance provided by sparse signals. This imbalance can lead to suboptimal visual quality, particularly in scenarios requiring high-level semantic alignment over time. To address this issue, we propose a degradation-based training strategy designed to regulate the relative influence of dense control signals and encourage more balanced utilization of both modalities. This strategy weakens the dominance of the dense input through controlled perturbations at both the feature and data levels:

\textit{1) Feature-level degradation:}
During training, with probability $\alpha$, we randomly scale the latent representation of the dense control by a factor $\lambda$ sampled uniformly from $[0.05, 1]$. Accordingly, Equation~\ref{eqn_fuse} can be reformulated as:
\begin{align}
    \label{eqn_fuse_deg}
    \boldsymbol{z}^l = \mathcal{F}^l( \boldsymbol{z}^{l-1}) + \phi^l(\lambda \cdot \mathcal{F}_\text{D}^l(\boldsymbol{c}^{l-1}_\text{D}) + \mathcal{F}_\text{P}^l(\boldsymbol{c}^{l-1}_\text{P}) ),
\end{align}
This operation reduces the magnitude of the dense features, making the model more reliant on complementary information provided by the sparse modality. Over time, this encourages the network to learn a more balanced integration of both control sources.

\textit{2) Data-level degradation:}  
Given a dense control tensor $D \in \mathbb{R}^{B \times C \times H \times W}$, we apply degradation with probability $\beta$ using two techniques:
\textit{a) Random Scale Fusion:}  
A set of spatial scales $\{1, 1/2, \dots, 1/2^n\}$ is predefined. One scale is randomly excluded, and the remaining scales are used to generate downsampled versions of the input, which are then upsampled to the original resolution. Each version is assigned a random weight normalized to sum to 1. Their weighted sum forms a fused depth map with multi-scale degradation and randomness, enhancing robustness to spatial variation.
\textit{b) Adaptive Blur Augmentation:}
An average blur with a randomly chosen odd-sized kernel is applied to the dense input to reduce sharpness, limiting the model’s tendency to overfit to local depth details.

Together, these degradations mitigate the model’s over-reliance on dense signals and enhance its ability to integrate complementary cues from sparse modalities, leading to more stable and consistent long-term video generation. During this pretraining stage, only the parameters of the control branches and fusion layers $\phi^l$ are updated, while the backbone remains frozen. This design introduces strong yet flexible conditioning from both dense and sparse modalities, preserving the stability and prior knowledge of the pre-trained base model.

\subsection{Frame Degradation Training}
The first image used in long-form video generation is itself a predicted frame whose quality gradually deteriorates over time, making it notably inferior to the clean images seen during training. As a result, the model tends to accumulate visual degradation across frames. Moreover, due to the non-lossless encoding and decoding process of the VAE, repeated reconstruction further amplifies this decline in quality.

To mitigate the gap between training and inference, we introduce a simple yet effective strategy: during training, we intentionally degrade the first image to mimic the corrupted inputs encountered in long-term generation. Specifically, we design two complementary degradation mechanisms that together define the degradation operator $\mathcal{T}(\cdot)$:  
\textbf{(1) Encoding degradation:} We simulate VAE-induced corruption by repeatedly encoding and decoding the first image $K$ times through the VAE and storing the resulting degraded reconstructions.  
\textbf{(2) Generation degradation:} We simulate diffusion-based degradation by adding Gaussian noise to the latent representation of the first image at a random timestep $t$, followed by denoising through the diffusion model to obtain a restored version. Formally, the degradation process can be expressed as:
\begin{equation}
\mathcal{T}(I) =
\begin{cases}
(\mathcal{D} \!\circ\! \mathcal{E})^{K}(I), & \textit{w.p. } 0.2,\\[4pt]
\mathcal{D}\!\big(\Phi_{0}(\sqrt{\alpha_t}\mathcal{E}(I){+}\sqrt{1-\alpha_t}\epsilon)\big), & \textit{w.p. } 0.8,
\end{cases}
\label{eq:degradation}
\end{equation}
where $I$ denotes the input image, $t<15$, and $\epsilon\!\sim\!\mathcal{N}(0,\mathbf{I})$.

\begin{figure}
\begin{center}
\includegraphics[width=\linewidth]{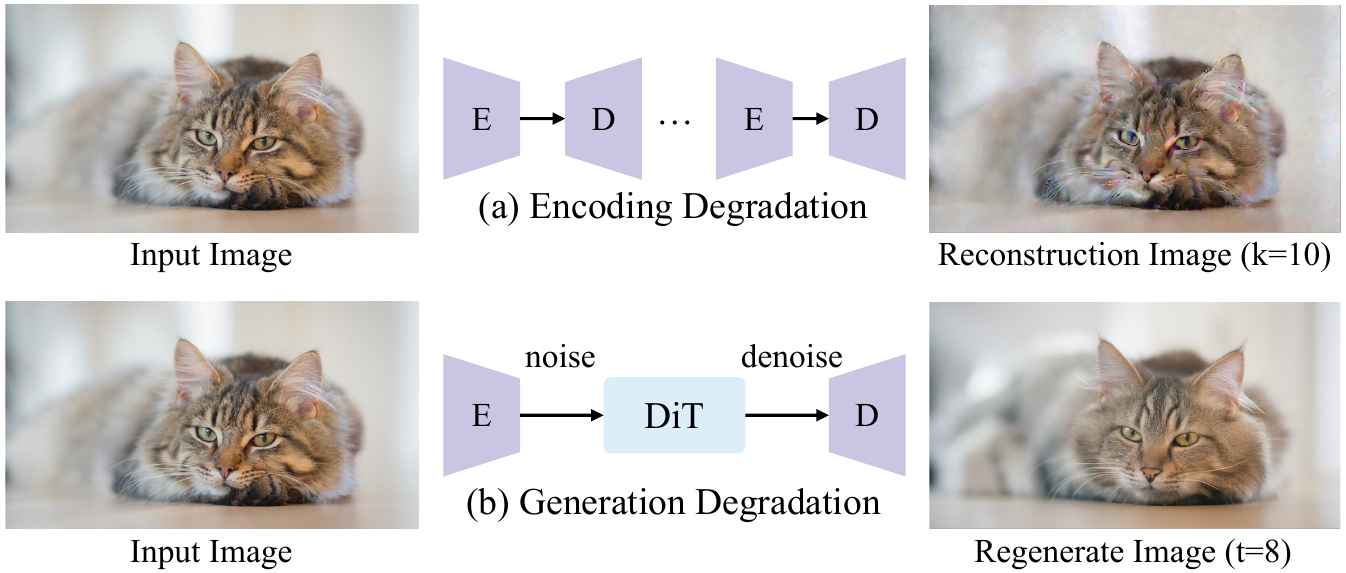}
\vskip -0.1in
\caption{
\textbf{Details of Frame Degradation.}
We apply two types of degradation—VAE encoding and denoising—to simulate the quality decay that occurs during long-term generation.
}
\label{fig:Image_degradation}
\end{center}
\vskip -0.2in
\end{figure}

\begin{figure*}
\begin{center}
\includegraphics[width=\linewidth]{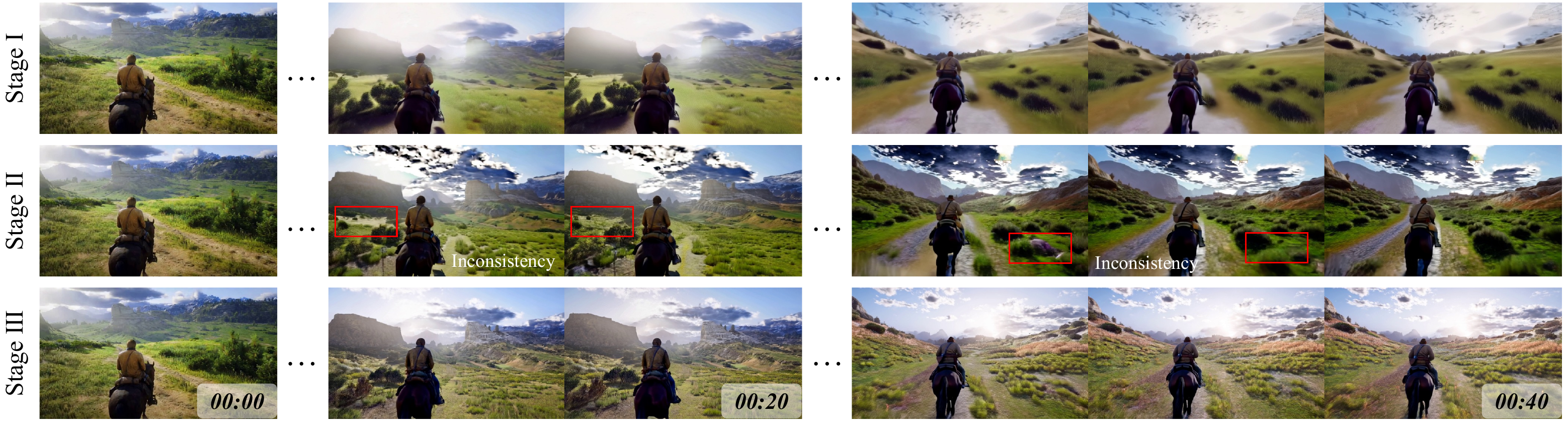}
\vskip -0.1in
\caption{\textbf{Visual comparison across training stages.} The results indicate that the second stage alleviates visual degradation but introduces intra-clip inconsistency, which is subsequently resolved by the final stage of training.}
\label{fig:ablation_degradation}
\end{center}
\vskip -0.2in
\end{figure*}

During training, we randomly apply $\mathcal{T}(\cdot)$ with a probability $\alpha$, where the degradation severity is inversely proportional to its sampling probability—milder degradations occur more frequently, while stronger degradations are applied less often. This strategy effectively improves the visual quality of long video generation.
However, despite the improved visual fidelity, this training scheme introduces a new issue of temporal inconsistency: the model tends to enhance the quality of the next clip independently, leading to noticeable discrepancies between adjacent clips.
To mitigate this problem, we introduce the history-context guidance stage, which explicitly incorporates the preceding clip as contextual conditioning to enforce temporal coherence across clips.

\subsection{History Context Guidance}
To enhance temporal consistency between adjacent clips and further improve long-term coherence, we propose \textbf{History Context Guidance}. During training, we introduce history frames to encourage the model to learn causal temporal dependencies. Specifically, we first use the VAE encoder $\mathcal{E}(\cdot)$ to obtain the latent representations of the $N_H$ history frames $V_H$, denoted as $\boldsymbol{z}_{H} = \mathcal{E}(V_H)$. Following the masking mechanism in Wan, we apply an all-ones mask to the history latent $\boldsymbol{z}_{H}$, forcing the model to reference historical context for maintaining temporal consistency.
To align the training and long-term inference domains, we apply the degradation operator $\mathcal{T}(\cdot)$ to the history frames, producing degraded inputs $\tilde{V}_H = \mathcal{T}(V_H)$. The degraded frames are then encoded as:
$
\tilde{\boldsymbol{z}}_H = \mathcal{E}(\tilde{V}_H).
$
Our model can be formulated as:
\begin{equation}
\boldsymbol{z}_t = \mathcal{F}(\boldsymbol{z}_{t+1} \mid \boldsymbol{z}_I, \boldsymbol{z}_H, \boldsymbol{c}_{D}, \boldsymbol{c}_{P}),
\end{equation}
where $\boldsymbol{z}_I$ denotes the latent representation of the initial frame, and $\boldsymbol{c}_{D}$ and $\boldsymbol{c}_{P}$ represent the dense and sparse control signals, respectively. Here, $\mathcal{F}$ denotes the overall generative model.

In clip-based long video generation, the first frame of each clip serves as the temporal anchor linking consecutive segments. However, its latent representation may deviate from both the historical context and the degraded input, potentially weakening cross-clip coherence. To alleviate discontinuities at this boundary frame, we assign exponentially increasing weights $\{0.05,\,0.325,\,0.757\}$ to the latents of the first three generated frames, encouraging smoother early-frame transitions within each clip. In addition, we regularize the first latent of the predicted frames from complementary frequency perspectives to further stabilize this boundary frame. To jointly reinforce reconstruction fidelity and maintain smooth temporal boundaries, we introduce three loss terms as described below:

\textbf{(1) History context consistency.}
We encourage the first predicted latent to align with the last latent of history context:
\begin{equation}
\mathcal{L}_{\text{cons}} =
\left\|
\boldsymbol{z}_{H}^{-1} - \hat{\boldsymbol{z}}^{0}
\right\|^{2},
\end{equation}
where $\boldsymbol{z}_{H}^{-1}$ is the last latent of history segment and $\hat{\boldsymbol{z}}^{0}$ is the first predicted latent.

\textbf{(2) Degradation consistency.}
To maintain structural coherence with degraded inputs and stabilize clip boundaries, we impose low-frequency consistency:
\begin{equation}
\mathcal{L}_{\text{deg}} =
\left\|
\mathcal{F}_{\text{lp}}\!\left(\tilde{\boldsymbol{z}}_{I}^{0}\right)
-
\mathcal{F}_{\text{lp}}\!\left(\hat{\boldsymbol{z}}^{0}\right)
\right\|^{2},
\end{equation}
where $\tilde{\boldsymbol{z}}_{I}^{0}$ is the latent of the degraded first image and $\mathcal{F}_{\text{lp}}(\cdot)$ denotes a low-frequency extraction operator.

\textbf{(3) Ground-truth high-frequency alignment.}
To preserve fine-grained appearance details, we align high-frequency components with the ground truth:
\begin{equation}
\mathcal{L}_{\text{gt}} =
\left\|
\mathcal{F}_{\text{hp}}\!\left(\boldsymbol{z}_{\mathrm{gt}}^{0}\right)
-
\mathcal{F}_{\text{hp}}\!\left(\hat{\boldsymbol{z}}^{0}\right)
\right\|^{2},
\end{equation}
where $\mathcal{F}_{\text{hp}}(\cdot)$ denotes a high-frequency operator.

The final temporal regularization objective is defined as:
\begin{equation}
\mathcal{L}_{\text{temp}} =
\lambda_{\text{deg}} \mathcal{L}_{\text{deg}}
+ \lambda_{\text{gt}} \mathcal{L}_{\text{gt}}
+ \lambda_{\text{cons}} \mathcal{L}_{\text{cons}}.
\end{equation}
In practice, we set $\lambda_{\text{deg}}$ = 0.2, $\lambda_{\text{gt}}$ = 0.15 and $\lambda_{\text{cons}}$ = 0.5. 
To further capture the causal dependencies of historical information, we additionally update all parameters in the self-attention layers of the base model. The number of history frames $N_H$ is uniformly sampled from $[0, 16]$, enabling LongVie~2 to perform unified long-video inference—whether or not historical references are available. This contextual conditioning substantially enhances temporal continuity and cross-clip consistency, resulting in smoother and more coherent long-video generation.

\subsection{Training-free Inter-clip Consistency Strategies}
To further improve temporal coherence across adjacent clips, we introduce two complementary \textbf{training-free} strategies that enhance inter-clip consistency.

\noindent \textbf{Unified Noise Initialization.}
To further enhance temporal stability and consistency during video generation, we adopt a unified noise initialization across all video clips. Rather than sampling a new noise latent for each clip, a single shared noise instance is maintained throughout the entire sequence, providing a coherent stochastic prior that strengthens temporal continuity and overall coherence.

\noindent \textbf{Global Normalization.}
Independent normalization of control inputs across clips often causes temporal discontinuities, as each segment may adopt a different depth scale. To mitigate this, we employ a global normalization strategy applied to the entire depth sequence. Specifically, we compute the 5th and 95th percentiles of all pixel values across the full video and use them as the global minimum and maximum bounds. The depth values are then clipped to this range and linearly scaled to $[0, 1]$. This percentile-based normalization is robust to outliers and ensures a consistent depth scale across clips. After normalization, the depth sequence is divided into overlapping clips to match the autoregressive inference process and enable point-map extraction.

\begin{table*}
\small
\centering
\setlength{\tabcolsep}{3mm}{
\caption{
\textbf{Quantitative comparison of LongVie~2 and baselines on LongVGenBench.} 
We compare \textbf{LongVie~2} with base diffusion models, controllable generation models, and world models on \textbf{LongVGenBench}, evaluating \textit{visual quality}, \textit{controllability}, and \textit{long-term consistency}. 
\textbf{Bold} indicates the best performance, and \underline{underline} denotes the second-best.
\label{tab:metrics}
}
\vskip -0.1in
{\begin{tabular}{lcccccccc}
\toprule
\multirow{2}{*}{\textsc{Methods}} 
& \multicolumn{2}{c}{\textbf{Quality}} 
& \multicolumn{2}{c}{\textbf{Controllability}} 
& \multicolumn{4}{c}{\textbf{Temporal Consistency}} \\
\cmidrule(lr){2-3} \cmidrule(lr){4-5} \cmidrule(lr){6-9}
 & A.Q.$\uparrow$ & I.Q.$\uparrow$ & SSIM$\uparrow$ & LPIPS$\downarrow$ & S.C.$\uparrow$ & B.C.$\uparrow$ & O.C.$\uparrow$ & D.D.$\uparrow$ \\
\midrule
\midrule
\rowcolor{gray!8}
Wan2.1~\cite{wan2025} & 49.72\% & 63.78\% & 0.406 & 0.488 & 83.56\% & 88.86\% & 20.30\% & 15.15\% \\
\midrule
VideoComposer~\cite{wang2023videocomposer} & 43.53\% & 59.33\% & 0.346 & 0.583 & 80.33\% & 88.30\% & 19.83\% & 27.78\% \\
Motion-I2V~\cite{shi2024motion} & 48.34\% & 61.57\% & 0.385 & 0.504 & 82.35\% & 89.25\% & 20.46\% & 44.13\% \\
Go-With-The-Flow~\cite{burgert2025gowiththeflow} & 53.59\% & 62.21\% & 0.453 & 0.394 & 84.37\% & 90.62\% & \underline{21.79\%} & 46.15\% \\
DiffusionAsShader~\cite{gu2025das} & 53.28\% & 64.57\% & 0.401 & 0.482 & \underline{86.06\%} & \underline{90.78\%} & 21.10\% & 36.76\% \\
\midrule
Matrix-Game-2.0~\cite{he2025matrix} & 55.24\% & 59.23\% & 0.427 & 0.501 & 77.82\% & 76.43\% & 19.37\% & 74.45\% \\
HunyuanGameCraft~\cite{li2025hunyuangamecraft} & \underline{56.18\%} & \underline{67.73\%} & \underline{0.483} & \underline{0.386} & 79.12\% & 74.24\% & 20.94\% & \underline{80.46\%} \\
\rowcolor{cyan!15}
\textbf{LongVie~2 (Ours)} & \textbf{58.47\%} & \textbf{69.77\%} & \textbf{0.529} & \textbf{0.295} & \textbf{91.05\%} & \textbf{92.45\%} & \textbf{23.37\%} & \textbf{82.95\%} \\
\midrule
\bottomrule
\end{tabular}}}
\vskip -0.05in
\end{table*}

\begin{figure*}
\begin{center}
\includegraphics[width=\linewidth]{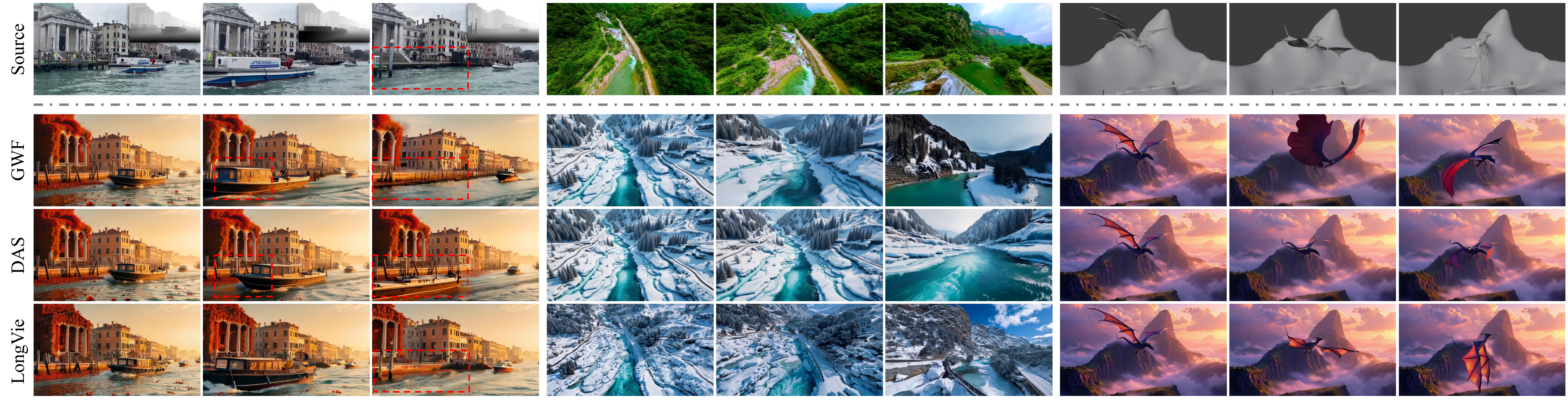}
\vskip -0.1in
\caption{
\textbf{Controllability comparison between LongVie~2 and baselines.}
\textbf{LongVie~2} demonstrates markedly superior controllability, maintaining precise structural alignment and realistic appearance across frames compared with the baselines.
\textit{GWF} denotes \textit{Go-With-The-Flow}, and \textit{DAS} denotes \textit{Diffusion as Shader}.
}
\label{fig:Controllability_comparison}
\end{center}
\vskip -0.1in
\end{figure*}

\begin{figure*}
    \centering
    \includegraphics[width=0.99\linewidth]{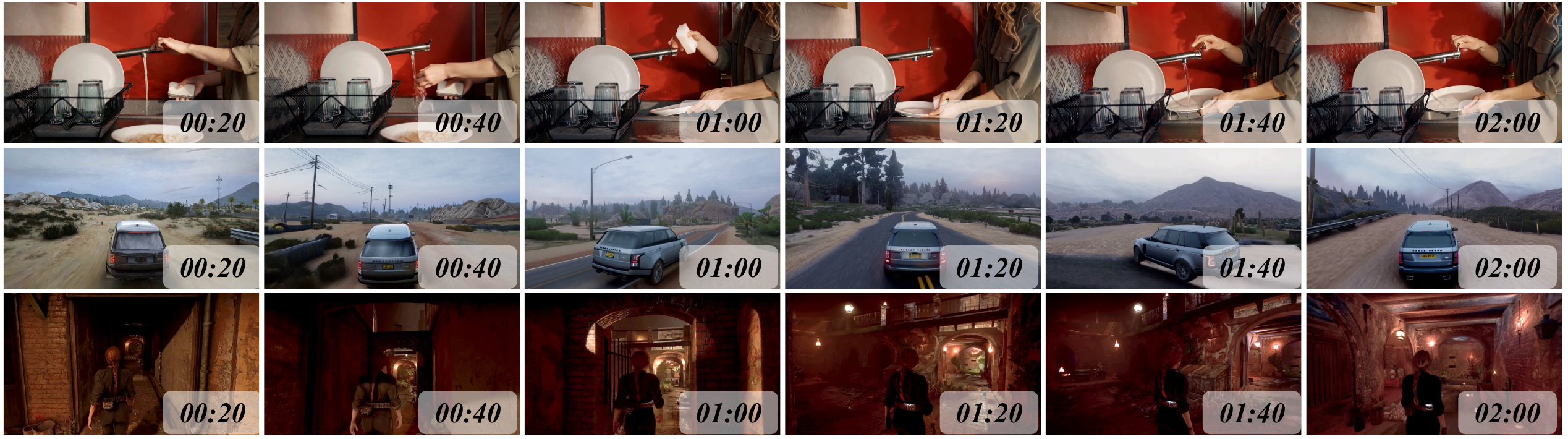}
    \vskip -0.1in
    \caption{
\textbf{Long-term generation demos.} We present several videos generated by \textbf{LongVie~2}, each lasting over two minutes, demonstrating sustained visual quality and long-term temporal consistency, further validating the effectiveness of our training strategy.
    }
    \label{fig:long_video_demo}
    \vskip -0.05in
\end{figure*}

\section{Experiments}
\noindent \textbf{Implementation Details.}
\label{experiments}
The video diffusion backbone of LongVie~2 is Wan2.1-I2V-14B~\cite{wan2025}. We replicate the first 12 DiT blocks to construct a dedicated control branch and further split each copied block along the feature dimension to form dense and sparse control pathways, effectively reducing computational overhead. During training, we first extract depth maps using Video Depth Anything~\cite{video_depth_anything} as dense control signals, and then apply SpatialTracker~\cite{SpatialTracker} to estimate and track 3D points from the normalized depth. Following DAS~\cite{gu2025das}, we uniformly sample 4,900 points per clip as sparse control inputs.
In total, around 100000 videos are used for training. In the first and second stages, approximately 60000 videos from ACID~\cite{acid}, Vchitect-T2V-DataVerse~\cite{dataverse}, and MovieNet~\cite{huang2020movienet} are employed. Each video is divided into 81-frame clips with a resolution of $352\times640$ at 16 fps to save memory and improve efficiency. LongVie~2 is optimized with AdamW at a learning rate of $1\times10^{-5}$, updating only the control branch parameters.
In the final stage, longer videos are incorporated to enable temporal reasoning over extended sequences. About 40000 videos from OmniWorld~\cite{zhou2025omniworld} and SpatialVID~\cite{wang2025spatialvid} are used, all downsampled to $352\times640$ at 16~fps. This stage employs AdamW with a learning rate of $5\times10^{-6}$ and updates the parameters in both the self-attention layers and the control modules.
All three stages are conducted on 16 A100 GPUs with each GPU using a batch size of 1, for approximately 5000, 1000, and 2000 iterations, respectively. The full training process takes about two days.

\noindent \textbf{LongVGenBench.}
To address the absence of suitable benchmarks for controllable long video generation, we introduce LongVGenBench—a dataset of 100 one-shot videos, each exceeding one minute in duration at 1080p resolution. Existing datasets fall short in this regard, as they lack long, continuous, one-shot videos that are essential for evaluating temporal consistency and controllability.
LongVGenBench covers diverse real-world and game-based scenarios and includes challenging cases such as rapid scene transitions and complex object motions (see the supplementary material for details), establishing it as a comprehensive and demanding benchmark for long video generation.
For evaluation, each video is divided into 81-frame clips at 16~fps with a one-frame overlap, following the autoregressive setup used in our experiments. Captions are automatically generated by Qwen2.5-VL-7B~\cite{Qwen2.5-VL} to serve as text prompts, and corresponding control signals are extracted from each clip. During validation, the first frame of each video is left unaltered to ensure fair comparison and to enable accurate assessment of generation quality against ground-truth references.

\subsection{Qualitative Comparison}
To comprehensively and fairly evaluate the performance of \textbf{LongVie~2}, we employ both objective and subjective metrics for comparison with existing baselines.

\noindent \textbf{Objective Evaluation.}
We first evaluate the effectiveness of LongVie~2 by comparing it against three categories of baselines:
\textit{(1) Pretrained video models}: Wan2.1~\cite{yang2024cogvideox};
\textit{(2) Controllable models}: VideoComposer~\cite{wang2023videocomposer}, Go-with-the-Flow~\cite{burgert2025gowiththeflow}, DAS~\cite{gu2025das}, and Motion-I2V~\cite{shi2024motion};
\textit{(3) World models}: Hunyuan-GameCraft~\cite{li2025hunyuangamecraft} and Matrix-Game~\cite{he2025matrix}.
Following the widely adopted VBench protocol~\cite{vbench}, we evaluate performance using seven metrics—\textit{Background Consistency (B.C.)}, \textit{Subject Consistency (S.C.)}, 
\textit{Overall Consistency(Q.C.)}, \textit{Dynamic Degree (D.D.)}, \textit{Aesthetic Quality (A.Q.)}, and \textit{Imaging Quality (I.Q.)}—to assess temporal coherence and visual fidelity. 
We also report similarity-based metrics, including SSIM and LPIPS, to measure the controllability of the generated videos.

As shown in Tab.~\ref{tab:metrics}, LongVie~2 achieves substantial improvements over controllable baselines in controllability, and surpasses existing other baselines in both temporal consistency and visual quality. 
These results demonstrate that LongVie~2 effectively maintains high controllability while producing long videos of superior visual coherence and quality, achieving state-of-the-art performance.

\begin{table}
\centering
\small
\caption{\textbf{Human Evaluation Results.} Sixty participants rated LongVie~2 and baseline models across five evaluation dimensions. LongVie~2 consistently achieved the highest scores in all aspects.}
\label{tab:userstudy}
\vskip -0.1in
\resizebox{\linewidth}{!}{
\begin{tabular}{lccccc}
\toprule
\textsc{Method} & VQ & PVC & CC & ColC & TC \\
\midrule
Matrix-Game-2.0~\cite{he2025matrix} & 2.12 & 2.08 & 2.27 & 2.23 & 2.37 \\
Go-With-The-Flow~\cite{burgert2025gowiththeflow} & 2.34 & 2.32 & 2.18 & 2.26 & 2.02 \\
DiffusionAsShader~\cite{gu2025das} & 3.03 & 3.07 & 3.09 & 3.02 & 2.79 \\
HunyuanGameCraft~\cite{li2025hunyuangamecraft} & 3.11 & 3.14 & 3.28 & 3.37 & 3.35 \\
\rowcolor{cyan!15}
\textbf{LongVie (Ours)} & \textbf{4.40} & \textbf{4.39} & \textbf{4.18} & \textbf{4.12} & \textbf{4.53} \\
\bottomrule
\end{tabular}}
\vskip -0.05in
\end{table}

\noindent \textbf{Human Evaluation.}
To further assess the perceptual quality of long video generation, we conduct a carefully designed human evaluation study.
To mitigate participant fatigue, the study follows a balanced and randomized sampling protocol.
From all generated videos, we randomly select 80 samples, each paired with its corresponding prompt and control signals.
Participants evaluate five key aspects: \textit{Visual Quality (VQ)}, \textit{Prompt–Video Consistency (PVC)}, \textit{Condition Consistency (CC)}, \textit{Color Consistency (ColC)}, and \textit{Temporal Consistency (TC)}.
We compare five representative models—Go-With-The-Flow~\cite{burgert2025gowiththeflow}, DAS~\cite{gu2025das}, HunyuanGameCraft~\cite{li2025hunyuangamecraft}, Matrix-Game~\cite{he2025matrix}, and LongVie~2.
A total of 60 participants are recruited. For each evaluation aspect, they rank the models from best to worst, assigning scores from 5 to 1 accordingly.
The averaged results, summarized in Tab.~\ref{tab:userstudy}, show that LongVie~2 consistently attains the highest scores across all criteria, demonstrating its superior perceptual quality and controllability.

\begin{table*}
\small
\centering
\setlength{\tabcolsep}{2.5mm}{
\caption{
\textbf{Ablation on staged training of LongVie~2.}
We progressively introduce three stages: \textit{Control Learning}, \textit{Degradation Adaptation}, and \textit{History Context}. 
Each stage consistently enhances visual quality, controllability, and long-term consistency.
\label{tab:ablation_study}
}
\vskip -0.1in
\begin{tabular}{lcccccccc}
\toprule
\multirow{2}{*}{\textsc{Training Strategy}} 
& \multicolumn{2}{c}{\textbf{Quality}} 
& \multicolumn{2}{c}{\textbf{Controllability}} 
& \multicolumn{4}{c}{\textbf{Temporal Consistency}} \\
\cmidrule(lr){2-3} \cmidrule(lr){4-5} \cmidrule(lr){6-9}
 & A.Q.$\uparrow$ & I.Q.$\uparrow$ 
 & SSIM$\uparrow$ & LPIPS$\downarrow$ 
 & S.C.$\uparrow$ & B.C.$\uparrow$ & O.C.$\uparrow$ & D.D.$\uparrow$ \\
\midrule
Base Model
    & 49.72\% & 63.78\% & 0.406 & 0.488 & 83.56\% & 88.86\% & 20.30\% & 15.15\% \\
~ + Control Learning
    & 51.36\% & 64.32\% & 0.456 & 0.376 & 85.94\% & 90.42\% & 20.95\% & 73.62\% \\
~ + Degradation Adaptation              
    & 54.08\% & 66.21\% & 0.501 & 0.328 & 88.12\% & 91.05\% & 21.56\% & 76.12\% \\
    \rowcolor{cyan!15}
~ + History Context
    & \textbf{58.47\%} & \textbf{69.77\%} & \textbf{0.529} & \textbf{0.295} & \textbf{91.05\%} & \textbf{92.45\%} & \textbf{23.37\%} & \textbf{82.59\%} \\
\bottomrule
\end{tabular}}
\vskip -0.1in
\end{table*}

\begin{table}
\small
\centering
\setlength{\tabcolsep}{1mm}{
\caption{\textbf{Ablation study for our proposed components.} The \textcolor{cyan!15}{\rule{1em}{1.5ex}} block denotes experiments targeting temporal consistency, while the \textcolor{purple!15}{\rule{1em}{1.5ex}} block denotes those focusing on visual quality.}
\label{tab:ablation_noise_norm}
\vskip -0.08in
{\begin{tabular}{lcccc}
\toprule
{\textsc{Methods}} & A.Q. & I.Q. & S.C. & B.C. \\
\midrule
\midrule
Full model  &  
\cellcolor{purple!15}\textbf{58.47\%} & 
\cellcolor{purple!15}\textbf{69.77\%} & 
\cellcolor{cyan!15}\textbf{91.05\%} & 
\cellcolor{cyan!15}\textbf{92.45\%} \\
\midrule
~ \textit{w/o} Global Normalization  & 
57.73\% & 
69.56\% & 
\cellcolor{cyan!15}88.81\% & 
\cellcolor{cyan!15}91.41\% \\	
~ \textit{w/o} Unified Initial Noise &  
57.80\% & 
69.62\% & 
\cellcolor{cyan!15}88.73\% & 
\cellcolor{cyan!15}91.59\% \\
~ \textit{w/o} Both                 & 
57.03\% & 
68.89\% & 
\cellcolor{cyan!15}88.56\% & 
\cellcolor{cyan!15}91.37\% \\
\midrule
~ \textit{w/o} Feature Degradation &  
\cellcolor{purple!15}56.92\% &  
\cellcolor{purple!15}67.23\% & 
89.94\% & 
92.15\% \\
~ \textit{w/o} Data Degradation &  
\cellcolor{purple!15}57.11\% &  
\cellcolor{purple!15}67.84\% & 
89.71\% & 
92.08\% \\
~ \textit{w/o} Both &  
\cellcolor{purple!15}56.74\% &  
\cellcolor{purple!15}67.01\% & 
89.57\% & 
91.99\% \\   
\midrule
\bottomrule
\end{tabular}}}
\vskip -0.1in
\end{table}

\subsection{Quantitative Results}

To comprehensively demonstrate the performance of LongVie~2, we present results from two key aspects: \textit{controllability} and \textit{long-term quality \& consistency}.

\noindent \textbf{Controllability Comparison.}
To verify that LongVie~2 preserves controllability, we directly evaluate it on standard controllable video generation tasks and compare it with strong baselines, including DAS~\cite{gu2025das} and GWF~\cite{burgert2025gowiththeflow}.
As illustrated in Fig.~\ref{fig:Controllability_comparison}, our model consistently outperforms the baselines across three representative sub-tasks: motion transfer, style transfer, and mesh-to-video generation.
It can be observed that LongVie~2 achieves superior controllability and produces finer visual details under the same settings.

\noindent \textbf{Long-Term Generation Results.}
We further present long-duration generation results of LongVie~2 in Fig.~\ref{fig:long_video_demo}.
Each row depicts a video exceeding two minutes in length, demonstrating the model’s ability to maintain high visual fidelity and long-term temporal coherence.
Notably, under world-level guidance, the model can simulate realistic physical phenomena such as the flow change after turning a faucet on or off, highlighting its capability for world-consistent video synthesis and further demonstrating the effectiveness of LongVie~2.
Additional long-form results are provided in the supplementary material for a more comprehensive evaluation.

\subsection{Ablation Study}
We conduct ablation studies to verify the effectiveness of each component and training strategy adopted in our method.

\noindent \textbf{Stage-Wise Training.}
LongVie~2 is trained in three progressive stages, each designed to enhance different aspects of video generation. To examine the contribution of each stage, we systematically remove them and report the results in Tab.~\ref{tab:ablation_study}. The first stage yields a clear improvement in SSIM and LPIPS, indicating stronger controllability. The second stage further enhances visual quality through degradation-aware learning, while the third stage notably improves global temporal consistency by incorporating history-context guidance. These results confirm that our stage-wise training strategy is both necessary and effective.

\noindent \textbf{Unified Initial Noise \& Global Normalization.}
We further analyze the influence of unified initial noise and global normalization of control signals on temporal stability and controllability. Videos are generated under three configurations: (1) without Global Normalization, (2) without Unified Initial Noise, and (3) without both. As shown in Tab.~\ref{tab:ablation_noise_norm}, removing either component leads to performance drops across all metrics, indicating that unified noise initialization and global normalization are both beneficial for consistent and high-quality controllable long video generation.

\section{Related Works}
\noindent \textbf{Controllable Video Generation.}
Recent advances in controllable video generation have enabled the synthesis of visually compelling and semantically aligned videos under diverse structural and motion constraints. VideoComposer~\cite{wang2023videocomposer} enhances controllability through multiple conditioning signals. Following the ControlNet~\cite{controlnet} paradigm, SparseCtrl~\cite{guo2024sparsectrl} introduces sparse structural guidance for fine-grained control, while Go-With-The-Flow~\cite{burgert2025gowiththeflow} and MotionI2V~\cite{shi2024motion} utilize optical flow to direct temporal motion. DAS~\cite{gu2025das} further employs 3D point maps for precise spatial and motion control. More recently, Cosmos-Transfer-1~\cite{cosmostransfer1} and LongVie~\cite{gao2025longvie} integrate multi-modal control signals to jointly improve visual fidelity and alignment in controllable video synthesis.

\noindent \textbf{World Models.}
Recent advances in video world modeling are powered by large-scale transformer and diffusion architectures that generate high-fidelity, temporally coherent videos conditioned on actions or camera trajectories. A series of world models~\cite{yan,yu2025gamefactory,genie3,li2025hunyuangamecraft,mao2025yume,zhang2025matrix,he2025matrix,hong2025relic,zhu2025astrageneralinteractiveworld,chen2025learning} follow this paradigm, focusing on camera-control-based simulation of dynamic environments. Leveraging large-scale video or gameplay data, these systems learn controllable and physically plausible dynamics, marking an important step toward general-purpose video world models capable of synthesizing consistent and interactive scenes under explicit control.

\noindent \textbf{Long Video Generation.}
Generating long, coherent videos remains a fundamental challenge due to accumulated temporal drift and quality degradation over extended horizons. StreamingT2V~\cite{StreamingT2V} adopts an autoregressive strategy to progressively extend video length. Diffusion Forcing~\cite{chen2024diffusionforcing, song2025historyguidedvideodiffusion, yin2025causvid} introduces causal constraints to improve temporal stability, while TTT~\cite{TTT}, LCT~\cite{LCT}, and MoC~\cite{cai2025mixture} explicitly incorporate long-range contextual dependencies to enhance visual and temporal consistency.
More recently, SVI~\cite{li2025stable} proposes an error-recycling fine-tuning paradigm that exposes the model to its own autoregressive outputs, thereby mitigating the mismatch between short-clip training and long-form inference. In parallel, Self-Forcing~\cite{huang2025selfforcing} leverages video-level distribution-matching objectives~\cite{yin2024improved, yin2024onestep}, together with related extensions~\cite{cui2025self++, yang2025longliverealtimeinteractivelong, yu2025videossm, lu2025reward, liu2025rolling, yi2025deep, zhuang2025flashvsr, yesiltepe2025infinity, zhang2025blockvidblockdiffusionhighquality}, to achieve long-horizon stability through autoregressive rollout and self-consistent training.
Inspired by the growing emphasis on narrowing the train-test discrepancy, we introduce a simple yet effective approach that simulates long-form degradation directly on the input frames during training, enabling the model to better preserve visual fidelity and temporal coherence over extended durations.

\section{Conclusion}
In this work, we explore how to extend pretrained video diffusion models toward building general-purpose video world models capable of generating coherent, controllable, and temporally consistent long videos. 
We empirically demonstrate that strengthening controllability before enhancing long-term visual quality is an effective and practical strategy, and we propose \textbf{LongVie~2}, a three-stage autoregressive framework that progressively improves controllability, visual fidelity, and long-term consistency.
First, we integrate dense and sparse controls to inject world-level guidance, providing complementary structural and semantic cues.
Second, a degradation-aware training strategy applied to first-frame inputs bridges the gap between training and long-horizon inference.
Finally, history context guidance models long-range dependencies to maintain temporal coherence across extended sequences.
To enable comprehensive evaluation, we establish \textbf{LongVGenBench}, a benchmark containing 100 one-minute videos covering diverse real-world and synthetic environments.
Extensive experiments demonstrate that LongVie~2 achieves state-of-the-art performance in controllable long video generation and can autoregressively synthesize high-quality videos lasting up to 3–5 minutes, marking a significant step toward video world modeling.

{
    \small
    \bibliographystyle{ieeenat_fullname}
    \bibliography{main}
}

\input{sec/X_suppl}

\end{document}

%% file: sec/X_suppl.tex
\clearpage
\setcounter{page}{1}
\maketitlesupplementary

\section{Overview of Supplementary Material}

This supplementary document provides additional technical details, datasets, and qualitative analyses to complement the main paper.
In Sec.~\ref{Implementation}, we elaborate on the implementation details of \textbf{LongVie}, including the training data, inference adaptation strategy, and model configuration.
In Sec.~\ref{LongVGenBench}, we offer a comprehensive description of the \textbf{LongVGenBench} dataset used for evaluating long-term consistency and controllability.
In Sec.~\ref{Ablation}, we conduct ablation studies on the key components of our frame degradation training strategy.
In Sec.~\ref{more_res}, we provide additional qualitative results under diverse styles and scenarios, including ultra-long video cases.

\section{More Implementation Details}
\label{Implementation}
\subsection{Training Data}

\noindent \textbf{Data for Stage 1 \& Stage 2.}
As described in the \textit{Implementation Details} section of the main paper, Stages~1 and~2 are trained on approximately 60,000 videos sourced from three complementary datasets.
\textit{(1) ACID and ACID-Large}~\cite{acid} provide thousands of aerial drone videos capturing diverse coastlines, natural landscapes, and outdoor environments. The videos follow the same format as RealEstate10K~\cite{RealEstate10K}, ensuring consistent metadata and camera-trajectory structure.
\textit{(2) Vchitect\_T2V\_DataVerse}~\cite{dataverse} is a large-scale corpus of more than 14 million high-quality Internet videos, each paired with descriptive and fine-grained textual annotations, offering rich appearance and motion diversity for training.
\textit{(3) MovieNet}~\cite{huang2020movienet} contains 1,100 full-length movies spanning multiple decades, regions, and genres, providing complex scenes, camera motions, and human-centric activities.
To standardize temporal resolution and facilitate stable training, all videos from these sources are uniformly converted into 81-frame clips sampled at 16 fps.

\noindent \textbf{Data for Stage 3.}
Stage~3 focuses on long-horizon temporal modeling and therefore requires explicit history context inputs. To support this stage, we incorporate long-form videos from \textit{OmniWorld}~\cite{zhou2025omniworld} and \textit{SpatialVID}~\cite{wang2025spatialvid}, which contain extended sequences with coherent scene dynamics. For each video, we extract 81-frame target segments beginning from the 20th frame; these segments serve as the prediction targets for the model. The corresponding control signals and textual prompts are generated following the same pipeline as Stages~1 and~2.
During training, all frames preceding each 81-frame segment are loaded as the history context, enabling the model to learn long-range dependencies and temporal evolution. From the complete set of processed samples, we randomly select 40,000 segments to construct the Stage~3 training split.

\noindent \textbf{Data Pre-processing.}
During our experiments, we observed that the presence of scene transitions (\eg cuts or abrupt changes in viewpoint) in the training data can significantly undermine temporal consistency during long-horizon generation, often leading to undesired transitions in the synthesized videos. To alleviate this issue, we apply the PySceneDetect toolkit~\cite{castellano2024pyscenedetect} to automatically identify scene boundaries and segment each raw video into transition-free clips. Each resulting segment is uniformly sampled at 16 fps and truncated to 81 consecutive frames to form a stable and temporally coherent training unit for LongVie.
For each 81-frame clip, we compute a comprehensive set of control signals. Depth maps are extracted using Video Depth Anything~\cite{video_depth_anything}, and point trajectories are estimated with SpatialTracker~\cite{SpatialTracker}, leveraging both the depth information and RGB frames. In addition, we generate descriptive captions for every clip using Qwen-2.5-VL-7B~\cite{Qwen2.5-VL} to provide semantically rich and contextually aligned textual guidance.
After applying this unified pre-processing pipeline to all videos, we obtain a curated corpus of approximately 100,000 video–control signal pairs. This dataset serves as the complete training foundation for LongVie, enabling consistent learning across diverse visual domains and motion patterns.

\begin{figure}
\begin{center}
\includegraphics[width=\linewidth]{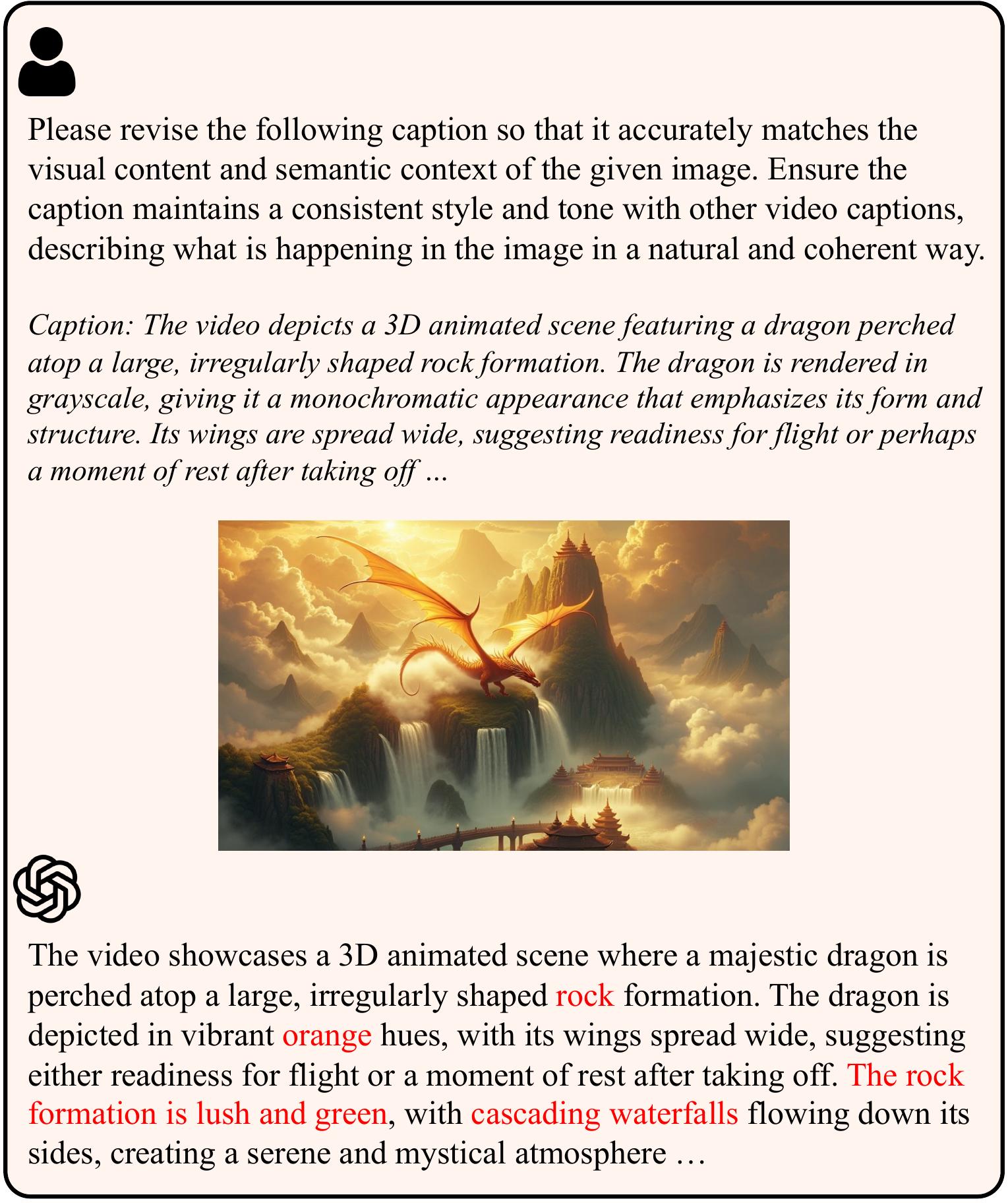}
\vskip -0.05in
\caption{
\textbf{Caption Refinement via MLLM.}
Given a new first-frame image and its original caption, we use Qwen-2.5-VL to refine the caption so that the prompt for video generation more accurately matches the visual content and remains stylistically consistent with the input image.
}
\label{fig:prompt_pipeline}
\end{center}
\vskip -0.1in
\end{figure}

\subsection{Inference Preparation.}
Point tracking naturally degrades during long-horizon inference because the tracked points depend on content visible in the first frame. As the video evolves—particularly when the originally tracked objects move out of view—the sparse tracks become unreliable. In our framework, point tracking acts as a sparse motion-control signal rather than an appearance cue. To preserve its effectiveness, we do not compute point tracks directly over the entire minute-long video. Instead, we first extract depth maps for the full sequence and apply global normalization across the entire duration. The video is then partitioned into overlapping 81-frame clips, and colorful point tracks are computed independently for each clip using the globally normalized depth, ensuring stable and consistent motion guidance throughout the generation process.
A similar challenge arises with captions. In transfer scenarios, the original captions of the source videos often become semantically misaligned with the visual content after transformation. To address this mismatch and remain consistent with our training pipeline, we apply Qwen-2.5-VL-7B~\cite{Qwen2.5-VL} to analyze the differences between the transferred frames and the original sequence. The model then revises the captions to more accurately reflect the updated visual semantics, as illustrated in Fig.~\ref{fig:prompt_pipeline}.

\subsection{Model Configuration.}
We provide additional implementation details of our LongVie model. During training, we apply \textit{input-frame degradation} with a probability of 20\%. Within this process, VAE-based degradation is selected 20\% of the time, while \textit{generation degradation} is used for the remaining 80\%. For encoding degradation, the parameter $k$ is sampled from [0, 10], with smaller values assigned higher sampling probability to simulate stronger degradation. For generation degradation, we select $t \in \{1, 5, 8, 10, 12\}$, again with a probability distribution favoring smaller values to emulate more severe degradation levels. In Stage~3, when history context is introduced, we apply the same degradation strategies to ensure consistent robustness across both current-frame and history-frame inputs.

In addition to frame-level degradation, we set the feature-level degradation probability $\alpha$ to 15\% and the data-level degradation probability $\beta$ to 10\%. At each training step, one or both degradation types are randomly activated. We adopt a fusion count of $n = 5$ for the \textit{Random Scale Fusion} module. To stabilize training, all degradation strategies are disabled during the first 2000 iterations and are gradually introduced during the final 1000 iterations.

For weight initialization, the dense and sparse branches within each MM Block inherit only half of the original weights from Wan2.1~\cite{wan2025}. Specifically, we interleave the pretrained weights by index—assigning weights at positions (0, 2, 4, $\dots$) to the dense branch and those at positions (1, 3, 5, $\dots$) to the sparse branch—while simultaneously halving the feature dimensionality. This “half-copy” initialization substantially reduces the total parameter count and provides a stable starting point for joint dense–sparse control learning.

\section{LongVGenBench}

\begin{figure*}
\begin{center}
\includegraphics[width=\linewidth]{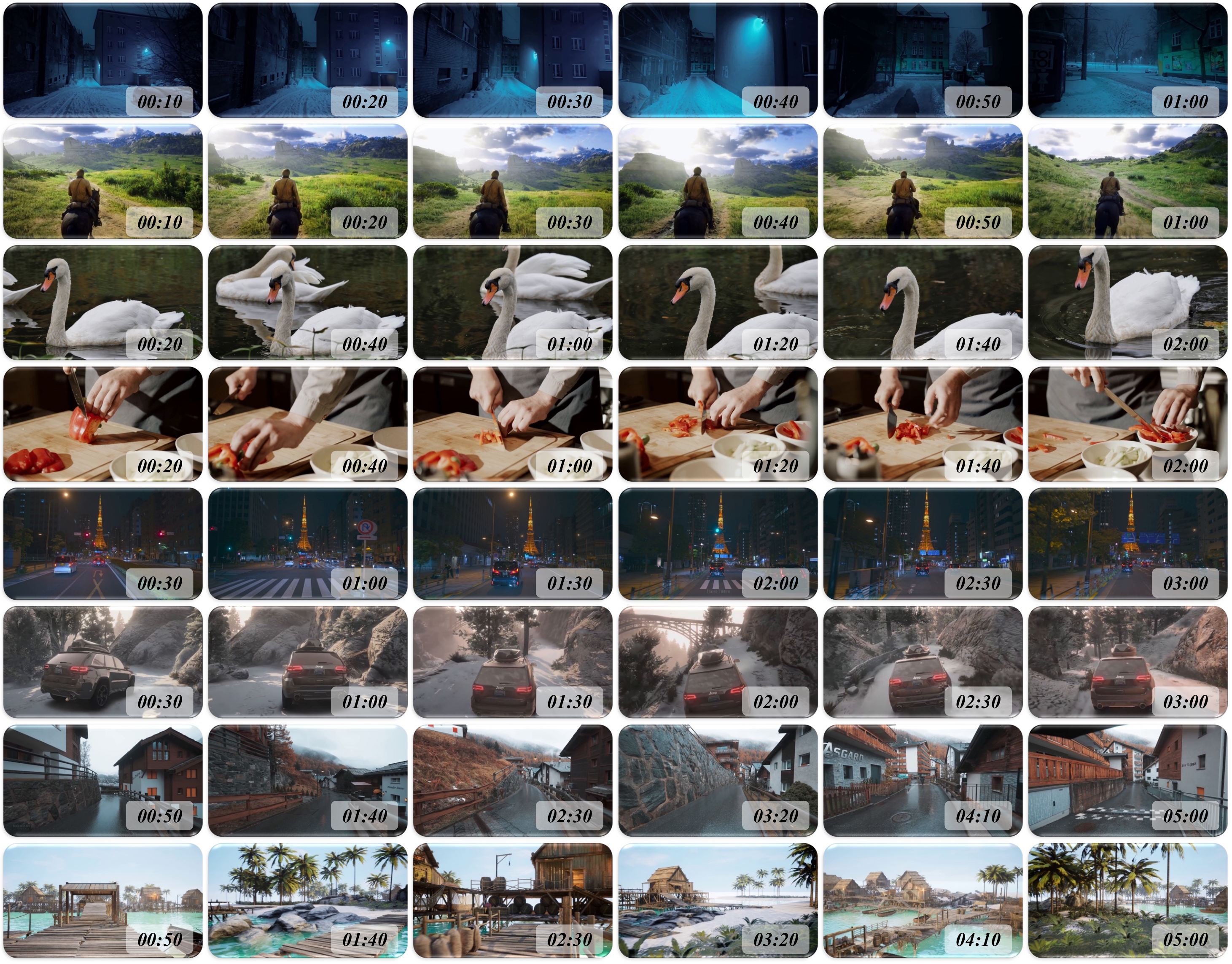}
\vskip -0.05in
\caption{\textbf{Examples from LongVGenBench.} We show several videos from both real-world and synthetic scenarios in LongVGenBench, covering a variety of indoor and outdoor environments to evaluate the controllable long video generation ability of our model.}
\label{fig:longbench}
\end{center}
\vskip -0.1in
\end{figure*}
\label{LongVGenBench}

\begin{table*}
\small
\centering
\setlength{\tabcolsep}{2.5mm}{
\caption{
\textbf{Ablation on the Degradation training of LongVie.}
We progressively introduce three stages: \textit{Control Learning}, \textit{Degradation Adaptation}, and \textit{History Context}. 
Each stage consistently enhances visual quality, controllability, and long-term consistency.
\label{tab:ablation_deg}
}
\vskip -0.05in
\begin{tabular}{lcccccccc}
\toprule
\multirow{2}{*}{\textsc{Training Strategy}} 
& \multicolumn{2}{c}{\textbf{Quality}} 
& \multicolumn{2}{c}{\textbf{Controllability}} 
& \multicolumn{4}{c}{\textbf{Temporal Consistency}} \\
\cmidrule(lr){2-3} \cmidrule(lr){4-5} \cmidrule(lr){6-9}
 & A.Q.$\uparrow$ & I.Q.$\uparrow$ 
 & SSIM$\uparrow$ & LPIPS$\downarrow$ 
 & S.C.$\uparrow$ & B.C.$\uparrow$ & O.C.$\uparrow$ & D.D.$\uparrow$ \\
\midrule
Base Model
    & 49.72\% & 63.78\% & 0.406 & 0.488 & 83.56\% & 88.86\% & 20.30\% & 15.15\% \\
~ + \textit{Encoding Degradation}
    & 52.18\% & 64.95\% & 0.451 & 0.402 & 85.72\% & 89.63\% & 20.66\% & 47.38\% \\
~ + \textit{Generation Degradation}
    & 53.96\% & 65.87\% & 0.478 & 0.356 & 87.64\% & 90.74\% & 21.14\% & 62.14\% \\
~ + \textit{Both}
    & \textbf{54.08\%} & \textbf{66.21\%} & \textbf{0.501} & \textbf{0.328} & \textbf{88.12\%} & \textbf{91.05\%} & \textbf{21.56\%} & \textbf{76.12\%} \\
\bottomrule
\end{tabular}}
\vskip -0.05in
\end{table*}

To better contextualize our proposed evaluation dataset, \textbf{LongVGenBench}, which is designed to assess both controllability and long-term quality and coherence in video world models, we present several representative examples in Fig.~\ref{fig:longbench}. The dataset consists of diverse real-world and synthetic scenes, each lasting at least one minute and captured at a minimum resolution of 1080p. 
As illustrated in the figure, LongVGenBench encompasses a wide range of camera motions, presenting substantial challenges for existing video generation models. Notably, the dataset is model-agnostic and can be used to evaluate any long-horizon video world model, not only LongVie.
In this paper, we utilize LongVGenBench by splitting each video into 81-frame clips with a one-frame overlap, followed by extracting captions and control signals for each clip. These short-clip captions are then used to construct the corresponding inference data.

\section{Additional Ablation Studies}
\label{Ablation}

\paragraph{Ablation for Degradation}
To further validate the effectiveness of our degradation strategy, we perform ablation studies on two degradation types: \textit{Encoding Degradation} and \textit{Generation Degradation}. The corresponding results are presented in Tab.~\ref{tab:ablation_deg}. These experiments demonstrate the contribution of each component to our overall frame degradation training pipeline and highlight their complementary effects.

\section{More Qualitative Results}
\label{more_res}
To further demonstrate the robustness and versatility of our LongVie model, we provide additional qualitative results across multiple long video generation scenarios, including 1-minute, 3-minute, and 5-minute cases. These examples highlight the model’s ability to maintain high visual quality, preserve global consistency, and adapt to diverse styles over extended temporal horizons.

\begin{figure*}[ht]
\begin{center}
\includegraphics[width=\linewidth]{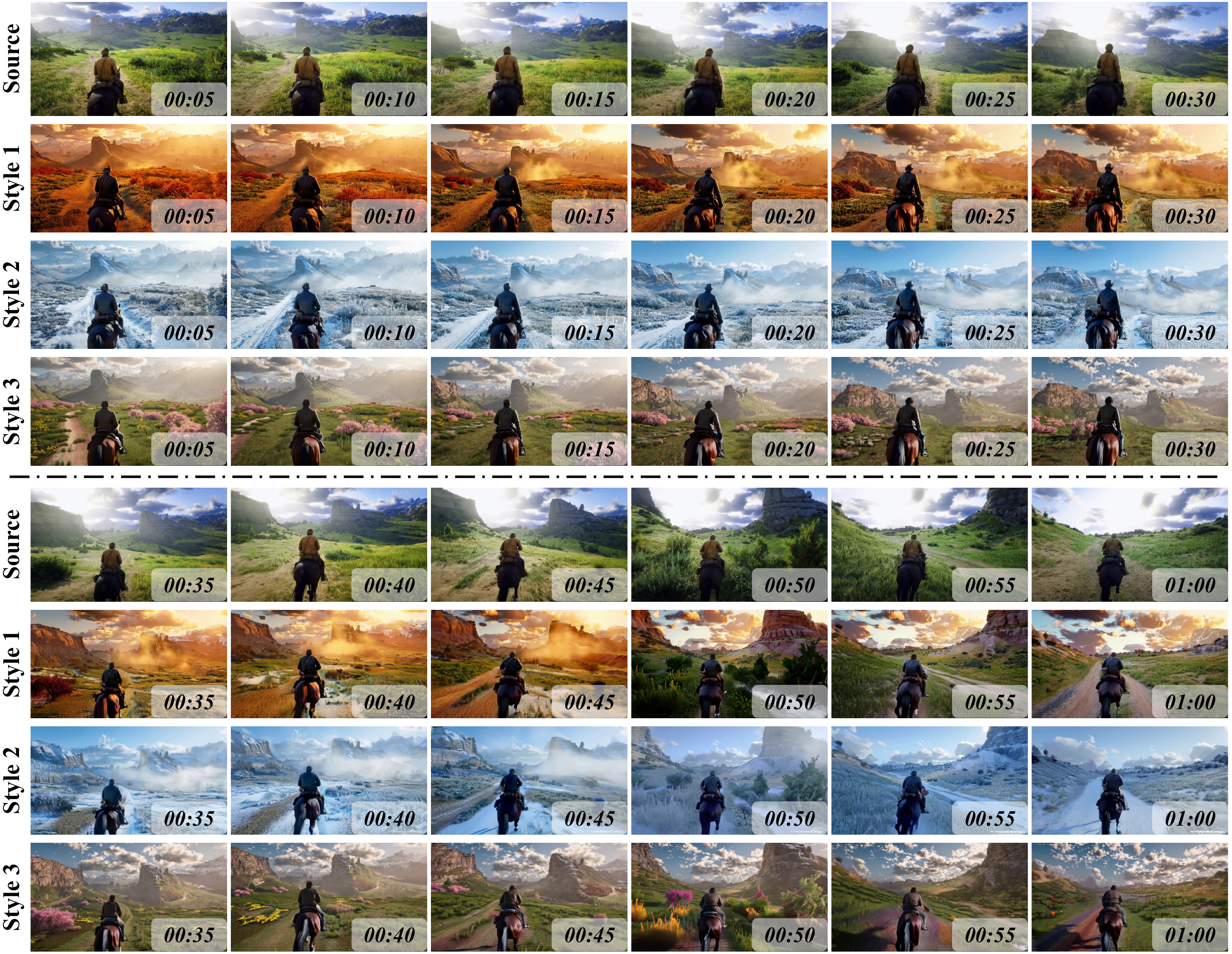}
\caption{\textbf{Subject-driven 1-minute scenario.} A man riding a horse, transferred into multiple seasonal styles while preserving motion dynamics and scene structure.}
\label{fig:suppl_transfer}
\end{center}
\end{figure*}

\begin{figure*}
\begin{center}
\includegraphics[width=\linewidth]{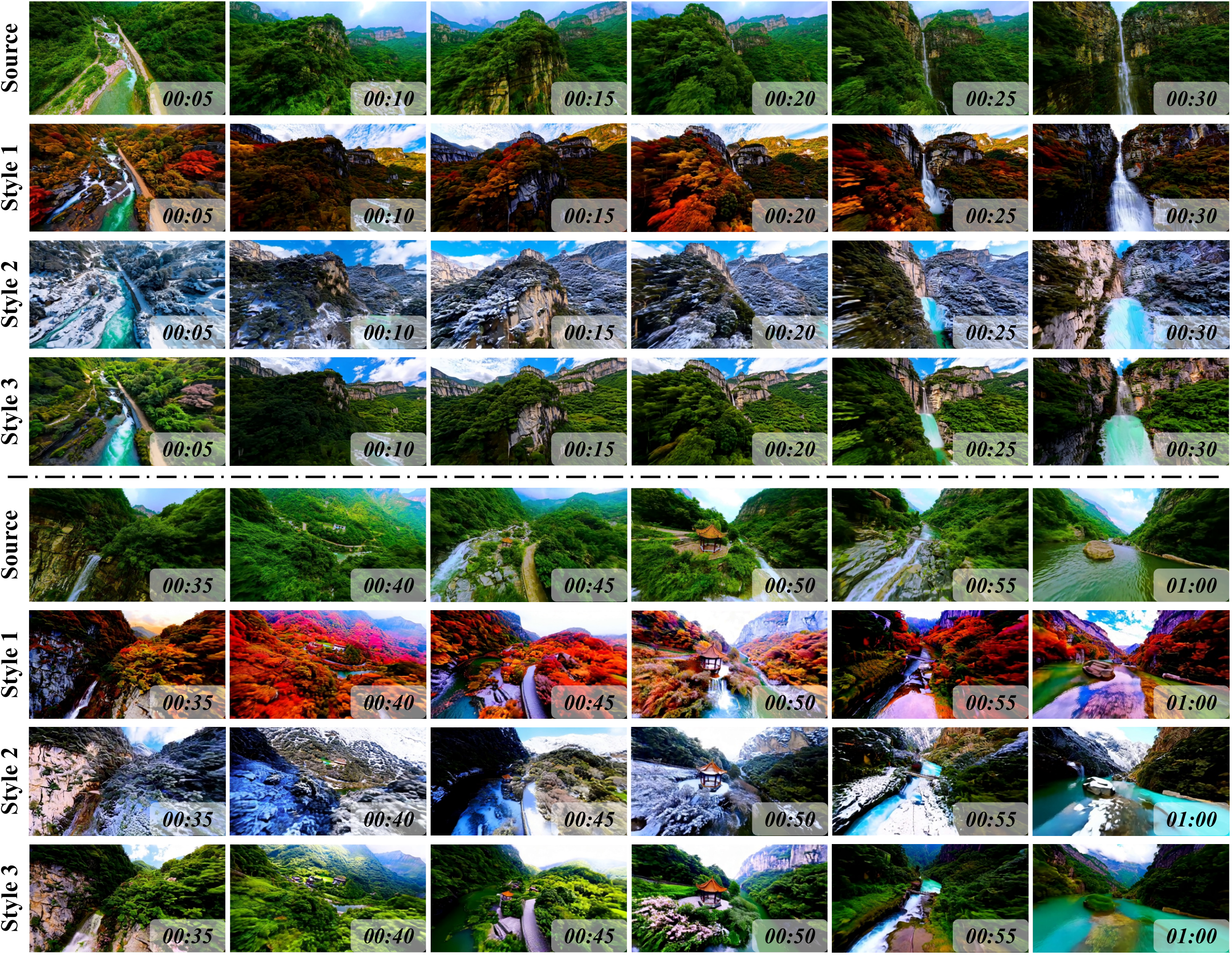}
\caption{\textbf{Subject-free scenario.} A drone-like car-drive sequence through a mountain valley, transferred across different seasonal styles with consistent global appearance.}
\label{fig:suppl_transfer2}
\end{center}
\end{figure*}

\noindent \textbf{1-minute Cases.}
For the 1-minute videos, we showcase two representative settings:
(1) a subject-driven scenario (a person riding a horse), and
(2) a subject-free scenario captured from a drone perspective.
Both settings are evaluated under three seasonal style transfers. As illustrated in Fig.~\ref{fig:suppl_transfer} and Fig.~\ref{fig:suppl_transfer2}, LongVie consistently preserves scene layout, motion dynamics, and semantic integrity while accurately adapting global textures, lighting, and seasonal atmospheres across different styles.

\begin{figure*}
\begin{center}
\includegraphics[width=\linewidth]{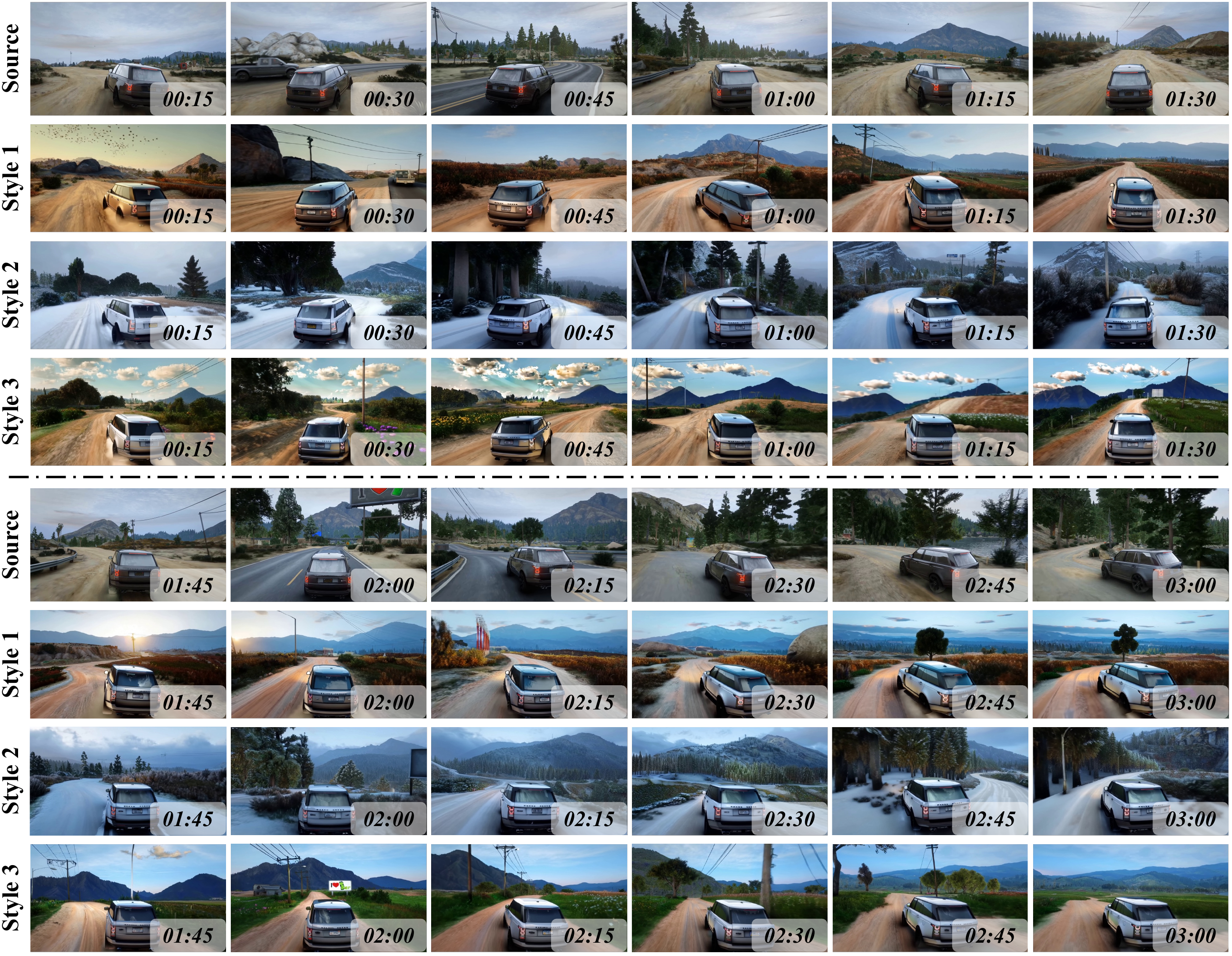}
\caption{\textbf{3-minute scenario.} A car driving through a mountain valley is transferred into multiple seasonal styles, demonstrating LongVie’s ability to maintain long-range temporal consistency and coherent global appearance over extended durations.}
\label{fig:suppl_transfer_3min}
\end{center}
\end{figure*}

\noindent \textbf{3-minute Cases.}
For the 3-minute scenarios, we evaluate long-term consistency under seasonal style variations. As shown in Fig.~\ref{fig:suppl_transfer_3min}, LongVie maintains coherent scene semantics and smooth transitions across all three seasonal styles.

\begin{figure*}
\begin{center}
\includegraphics[width=\linewidth]{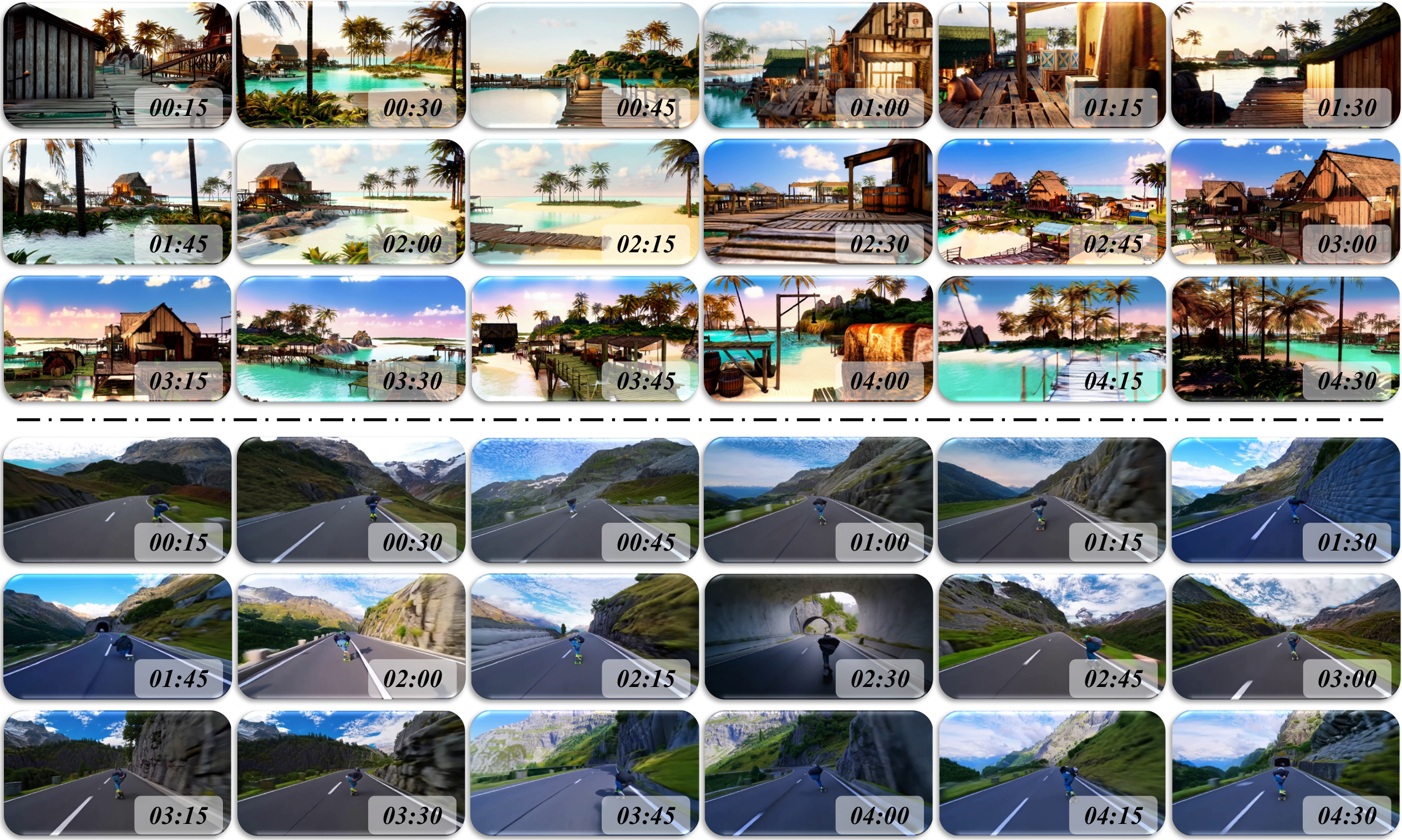}
\caption{\textbf{5-minute scenarios.} We present two ultra-long video cases: a subject-driven sequence and a subject-free sequence. These results illustrate LongVie’s robustness in preserving structural stability, motion coherence, and style consistency across 5-minute continuous generations.}
\label{fig:suppl_transfer_5min}
\end{center}
\end{figure*}

\noindent \textbf{5-minute Cases.}
To highlight LongVie’s capability in ultra-long video generation, we present two 5-minute examples covering both a subject-driven and a subject-free scenario. As shown in Fig.~\ref{fig:suppl_transfer_5min}, LongVie preserves global structure, maintains coherent motion behavior, and sustains stable visual appearance throughout the entire 5-minute duration.

\clearpage

\section{Limitations and Future Work}
LongVie presents a practical framework for adapting a pretrained video model into a controllable world model capable of generating coherent visual dynamics over 3–5 minutes. To systematically evaluate our training strategies while keeping computational costs tractable, all experiments in this paper are conducted at a resolution of 352$\times$640. Although this setting allows for comprehensive ablations and long-horizon testing, the relatively low resolution limits the model’s ability to express fine-grained details and high-frequency structures. In future work, we aim to scale LongVie to higher resolutions in order to further improve visual fidelity, capture richer scene dynamics, and more fully showcase the potential of LongVie as a video world model.